\documentclass[11pt]{article}

\usepackage[preprint]{acl}

\usepackage{times}
\usepackage{latexsym}

\usepackage[T1]{fontenc}

\usepackage[utf8]{inputenc}
\usepackage{graphicx}   
\usepackage{subcaption} 

\usepackage{microtype}

\usepackage{inconsolata}
\usepackage{amsmath,amssymb}

\usepackage{graphicx}
\usepackage{amsmath}
\usepackage{graphicx}
\usepackage{algorithm}
\usepackage{algorithmic}
\usepackage{tikz}
\usetikzlibrary{shapes, arrows, positioning, fit}
\usepackage{graphicx}   
\usepackage{booktabs}   
\usepackage{graphicx}    
\usepackage{booktabs}    
\usepackage{multirow}    
\usepackage{setspace}    
\usepackage{arydshln}
\usepackage{booktabs}
\usepackage{siunitx}
\usepackage{multirow}
\usepackage{xcolor,colortbl}
\usepackage[table]{xcolor}

\title{TrimTokenator-LC: Towards Adaptive Visual Token Pruning for Large Multimodal Models with Long Contexts}

\author{
\bf Hao Zhang\thanks{These authors contribute equally to this work.}\quad
Mengsi Lyu\footnotemark[1]\quad
Bo Huang\quad
Yulong Ao\thanks{Corresponding author.}\quad
Yonghua Lin\footnotemark[2] \\
Beijing Academy of Artificial Intelligence (BAAI)
}

\begin{document}
\maketitle
\begin{abstract}
Large Multimodal Models (LMMs) have proven effective on various tasks. They typically encode visual inputs into Original Model sequences of tokens, which are then concatenated with textual tokens and jointly processed by the language model. However, the growing number of visual tokens greatly increases inference cost. Visual token pruning has emerged as a promising solution. However, existing methods often overlook scenarios involving long context inputs with multiple images. In this paper, we analyze the challenges of visual token pruning in long context, multi-image settings and introduce an adaptive pruning method tailored for such scenarios. We decompose redundancy into intra-image and inter-image components and quantify them through intra-image diversity and inter-image variation, which jointly guide dynamic budget allocation. Our approach consists of two stages. The intra-image stage allocates each image a content-aware token budget and greedily selects its most representative tokens. The inter-image stage performs global diversity filtering to form a candidate pool and then applies a Pareto selection procedure that balances diversity with text alignment. Extensive experiments show that our approach can reduce up to 80\% of visual tokens while maintaining performance in long context settings.
\end{abstract}

\section{Introduction}
Large Multimodal Models (LMMs) \cite{bai2025qwen25vl,team2025kimivl,zhu2025internvl3,liu2024llava15,lin2023videollava} have substantially extended the inferential reach of Large Language Models (LLMs) \cite{brown2020language,touvron2023open,chiang2023vicuna,zhu2023a,li2023,zhang2023,huang2023,wang2023} by enabling unified reasoning over heterogeneous inputs that combine vision and language. Standard pipelines first convert images into Original Model visual token sequences via a vision encoder, after which these tokens are fused with textual tokens and processed jointly within a transformer backbone. Although this integration strategy supports strong multimodal understanding, it also produces exceptionally long sequences that often span several thousand tokens. The quadratic scaling behavior of attention with respect to sequence length \cite{vaswani2017attention,huggingface2024mastering,chen2023survey,keles2023computational,liu2022pyraformer} yields substantial memory and computational overhead under such conditions. These costs present nontrivial barriers to deploying LMMs in practice, particularly in settings where inference latency or computational resources are tightly constrained \cite{chen2024fastv,lin2025boosting}.
\begin{figure*}[t]
    \centering
    \includegraphics[width=1\textwidth]{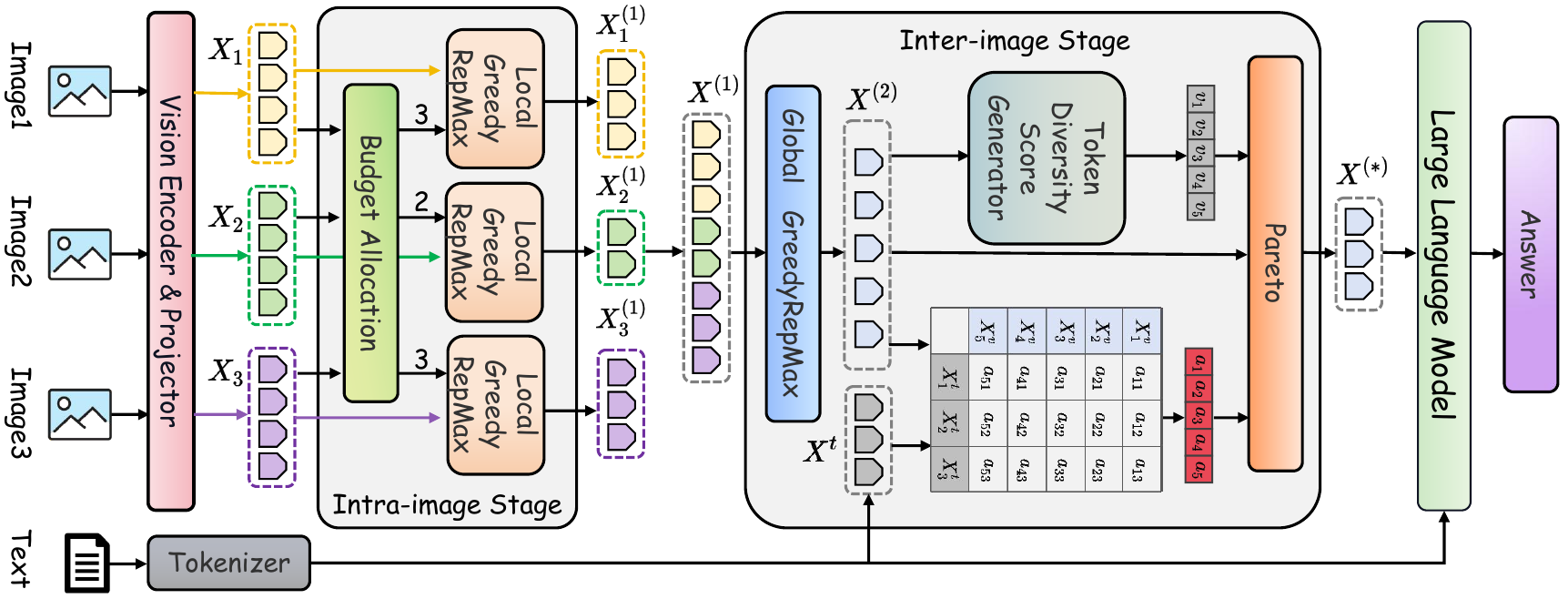}
    \caption{Comprehensive overview of the pruning method. We first perform an Intra-image Stage, where the visual token budget is adaptively allocated based on the diversity of each image. Within each image, a greedy representative selection strategy is applied to retain the most representative local visual tokens. Next, in the Inter-image Stage, greedy representative filtering is conducted over the candidate tokens collected from all images. Finally, by jointly considering token diversity and semantic alignment with the text, pareto selection is used to preserve tokens that are both informative across images and highly aligned with the text. Figure \ref{f_2} provides a brief description.}
    \label{f_1}
\end{figure*}

Recent studies indicate that visual tokens in LMMs is heavily overcomplete, with many visual embeddings contributing minimal semantic value to multimodal reasoning \cite{liu2024mustdrop,shang2024llava,huang2024ivtp,tong2025flowcut,li2025todre}. This observation has motivated a growing literature on strategies that selectively suppress redundant visual tokens during inference. Reducing the number of visual tokens shortens the multimodal sequence delivered to the transformer, which in turn mitigates the quadratic computational burden inherent in attention. A consistent finding across prior work is that substantial reductions in visual tokens can be achieved with only marginal impact on performance \cite{zhang2024fastervlm,chen2024fastv,lin2025boosting,huang2024ivtp,sunvelar}.

Existing visual token pruning methods can be broadly categorized into three types. (1) One prevalent direction leverages attention scores to identify redundant tokens and discard those with limited contribution \cite{lin2025boosting,shang2024llava,tong2025flowcut}. (2) Another class of methods introduces model-specific calibration or fine-tuning \cite{ye2025fitprune,lin2025boosting,li2025tokenpacker,cai2024matryoshka}, which incurs substantial computational cost and restricts scalability in practical applications. (3) Another line of work emphasizes representation diversity and semantic relevance \cite{zhang2025trimtokenator,zhang2025beyond, alvar2025divprune, li2025todre}, seeking to preserve tokens that carry the most informative content. However, all these methods largely overlook cases where the inputs involve long contexts. In such scenarios, the input sequence can be extremely long and often includes multiple images. Consequently, designing an effective token pruning strategy tailored to long context tasks becomes a considerably challenging problem.

In this paper, we analyze the challenges of visual token pruning in long context
settings. Accordingly, we propose an adaptive visual token pruning method designed for multi-image long context scenarios. We categorize redundancy into intra-image and inter-image forms, and model these two aspects using intra-image diversity  and inter-image variation. These measures guide the dynamic allocation of pruning budgets. Our method is composed of two procedures, intra-image pruning and inter-image pruning. The intra-image pruning procedure allocates different token budgets to each image based on its visual richness and employs a greedy strategy to select the most representative tokens, thereby yielding a compact representation for each image. The inter-image pruning procedure performs global diversity filtering over all image tokens to obtain a candidate set. It then applies Pareto selection based on diversity and text alignment indicators to retain the tokens that contribute most across images and are most relevant to the textual context. These tokens form the final visual representation. Extensive experiments demonstrate that our method preserves strong performance in long context scenarios while substantially reducing the number of visual tokens. In summary, our contributions can be summarized as follows:
\begin{itemize}
  \item We identify the key challenges of visual redundancy in long context settings and introduce an adaptive visual token pruning method for multi-image inputs, which allocates pruning budgets by modeling intra-image diversity and inter-image variation.
  \item We perform intra-image pruning by allocating per-image token budgets based on visual richness and using a greedy strategy to retain the most representative tokens, yielding compact visual representations.
  \item We conduct inter-image pruning by applying global diversity filtering over all image tokens to obtain a candidate set, and then use Pareto selection to retain the tokens that contribute most across images and most relevant to text.
\end{itemize}

\section{Related Work}
\textbf{Visual Token Pruning} aims to reduce computational cost and speed up inference by removing redundant or weakly informative visual tokens. \textbf{(1) One major line of research relies on attention statistics to estimate token importance.} PruMerge \cite{shang2024llava} merges clustered tokens in the vision encoder according to attention sparsity. FastV \cite{chen2024fastv} uses early attention signals to guide removal. SparseVLM \cite{zhang2024sparsevlm} performs text conditioned token selection through cross modal attention, and VisionZip \cite{yang2025visionzip} compresses visual inputs based on CLS attention in the final vision encoder layer. FlowCut \cite{tong2025flowcut} identifies redundancy by tracing information flow across attention layers, while LVPruning \cite{sun2025lvpruning} measures importance from interactions between visual and language tokens. \textbf{(2) Another direction uses calibration based practices, where pruning is determined by model behavior on held out data.} FitPrune \cite{ye2025fitprune} adjusts selection by comparing attention distributions before and after pruning. VTW \cite{lin2025boosting} locates layers after which token removal is reliably safe using calibration signals. \textbf{(3) A complementary perspective focuses on representativeness and semantic relevance.} TrimTokenator \cite{zhang2025trimtokenator} estimates cross modal alignment with visual text mutual information and incorporates visual diversity for adaptive pruning. DivPrune \cite{alvar2025divprune} retains a diverse subset by maximizing minimum pairwise token distance. ToDRE \cite{li2025todre} handles diversity and task relevance as independent factors, selecting representative tokens and later filtering task irrelevant ones during decoding. CDPruner \cite{zhang2025beyond} formulates the task as conditional diversity maximization and applies a determinantal point process to obtain tokens.

\section{Preliminaries and Challenges}
\subsection{Background of TrimTokenator}
TrimTokenator \cite{zhang2025trimtokenator} introduces two mutually reinforcing mechanisms that address visual token redundancy by jointly improving cross-modal alignment and greedily maximizing intra-modal representational coverage.

To ensure cross-modal semantic consistency, TrimTokenator estimates the value of each visual token by measuring its alignment with text tokens. Given the visual token set $\mathcal{X}^v = \{x_1^v, \ldots, x_N^v\}$ and the text token set $\mathcal{X}^t = \{x_1^t, \ldots, x_M^t\}$, the alignment score \(a_i\) for each visual token \(x_i^v\) is defined as follows:
\begin{equation}
a_i = - \frac{1}{M} \sum_{j=1}^{M} \left\lVert x_i^v - x_j^t \right\rVert_2^2 
\end{equation}
where smaller $a_i$ values indicate that the visual token $x_i^v$ is poorly aligned with textual semantics and should therefore be pruned. This scoring function facilitates the removal of semantically irrelevant regions, reducing visual noise and enhancing downstream text generation.

To enhance intra-modal representational diversity and reduce redundancy, TrimTokenator introduces a selection function \(GreedyRepMax(\cdot)\). Given a visual token set \(\mathcal{X}^v\) and a target budget \(N\), the function produces a subset \(\hat{\mathcal{X}}^v_N\) by greedily choosing tokens that maximize dispersion in the embedding space, thereby covering broader and more complementary visual semantics. This procedure approximates the following objective:
\begin{equation}
\hat{\mathcal{X}}^v_N = \arg\max_{\mathcal{X}\subseteq\mathcal{X}^v,\;|\mathcal{X}|=N} \mathbb{E}_{x_i,x_j\in\mathcal{X}} D(x_i,x_j)
\end{equation}
where \(D(x_i, x_j)\) denotes the pairwise token distance between \(x_i\) and \(x_j\). Guided by this objective, the selection can be expressed as follows:
\begin{equation}
\label{div}
\hat{\mathcal{X}}^v_N = GreedyRepMax(\mathcal{X}^v, N)
\end{equation}
which yields a more dispersed and less redundant token subset that better covers visual semantics.

\subsection{Key Challenges}
In long context multi-image scenarios, the model needs to encode visual tokens drawn from multiple images simultaneously. There are four key challenges: \textbf{(1) Intra-Image and Inter-Image Redundancy Modeling }Multi-image inputs exhibit two forms of redundancy. Intra-image redundancy arises from repeated textures or similar local structures within a single image, while inter-image redundancy emerges when different images share similar subjects or layouts. Characterizing these two types of redundancy remains a challenging problem. \textbf{(2) Cross-Image Pruning Budget Allocation }Images vary widely in visual complexity and semantic contribution. A fixed pruning ratio often leaves simple images with an excess of redundant tokens while compressing complex images too aggressively. Consequently, dynamically allocating token budgets based on image content becomes essential for preserving critical information. \textbf{(3) Sequence of Intra-Image and Inter-Image Pruning }Determining the order of these two stages is a key design decision. In this work, we prioritize intra-image pruning to first construct compact semantic bases for each image, ensuring that each image retains an independent and minimally sufficient representation. Placing inter-image pruning at the beginning would force all tokens to compete for a limited budget without constraints, which may result in some images being nearly pruned out, destroying their standalone semantic structure and causing their information to vanish from the overall representation. \textbf{(4) Position of Text Guided Alignment }Deciding whether to perform text guided alignment during the intra-image or inter-image pruning stage is an important consideration. Text alignment primarily targets global semantics rather than the local structure of individual images. Performing this during the inter-image pruning stage is more consistent with semantic reasoning, since textual descriptions typically capture relationships, key objects and trends across multiple images. The effectiveness of these design choices is verified in ablation studies.

\section{Methodology}
\subsection{Overview}
\begin{figure}[!t]
    \centering
    \includegraphics[width=1\linewidth]{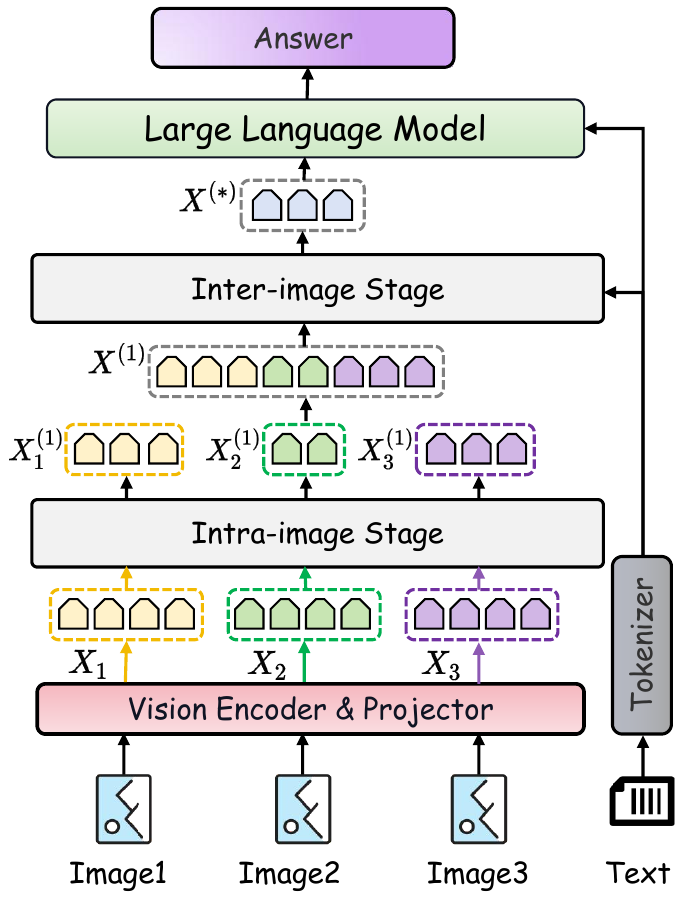}
    \caption{Brief overview of the visual token pruning method. Figure \ref{f_1} provides a detailed description.}
    \label{f_2}
\end{figure}

Given an image sequence $(I_1,\dots,I_n)$, the visual encoder produces for each image $I_k$ a set of tokens $X_k=\{x_{k,1},\dots,x_{k,m_k}\}$, where $m_k = |X_k|$ is the number of visual tokens for $I_k$. All tokens across the sequence can be represented as $X=\bigcup_{k=1}^n X_k$, with total size $M_0$. We aim to retain only $M_{final}$ tokens ($M_{final} < M_0$) such that the visual input remains both compact and comprehensive, while reducing computational cost. To achieve this, we perform intra-image pruning to compress the original token set \(X\) of each image, obtaining \(X^{(1)}\) and reducing the total number of tokens from \(M_0\) to \(M_1\). Then, we apply inter-image pruning to refine \(X^{(1)}\) by removing cross-image redundancies, ultimately producing the final token set \(X^{(*)}\) with \(M_{{final}}\) tokens. Figures \ref{f_1} and \ref{f_2} provide detailed and brief descriptions of our method, respectively.

Visual token redundancy in multi-image inputs arises from two structural sources.  \textbf{(1) Intra-image redundancy} occurs when different regions  within an image are semantically similar or structurally repetitive, generating many similar local tokens. \textbf{(2) Inter-image redundancy} arises when multiple images share repeated objects, scenes, or layouts. To adaptively control token retention, we introduce two metrics: \textbf{intra-image diversity ($\mathcal{D}_{intra}$)} and \textbf{inter-image variation ($\mathcal{D}_{inter}$)}, capturing both the internal complexity of a single image and the differences across images. We provide a detailed analysis of the metrics in Appendix \ref{app:metric}.

\paragraph{\boldmath$\mathcal{D}_{{intra}}$}
We define the intra-image diversity of an image $I_k$, denoted as $\mathcal{D}_{intra}(I_k)$, to measure the dispersion of visual tokens within the image, which is computed as follows:
\begin{equation}
\scalebox{0.85}{$
\displaystyle
\mathcal{D}_{intra}(I_k) = \frac{1}{m_k (m_k - 1)} 
\sum_{i \neq j} \left( 1 - \cos(x_{k,i}, x_{k,j}) \right)
$}
\end{equation}
Images with repetitive structures produce similar token representations, leading to lower $\mathcal{D}_{intra}(I_k)$, while images with richer content yield higher values. We further average this metric over all images to capture the overall intra-image complexity of the input set, which can be represented as follows:
\begin{equation}
\mathcal{D}_{intra} = \frac{1}{n} \sum_{k=1}^{n} \mathcal{D}_{intra}(I_k)
\end{equation}
Higher $\mathcal{D}_{intra}$ indicates richer internal visual content, suggesting that a higher token retention ratio should be used during intra-image pruning to preserve semantic structures.

\paragraph{\boldmath$\mathcal{D}_{{inter}}$}
To capture semantic changes across images, we measure each image relative to its predecessor using global semantic embeddings. Specifically, we aggregate all visual tokens of $I_k$ and $I_{k-1}$ into their respective global representations and compute the cosine distance between them, defined as follows:
\begin{equation}
\small
d_k = 1 - \cos\!\big( \frac{1}{m_k}\sum_{i=1}^{m_k} x_{k,i},\, \frac{1}{m_{k-1}}\sum_{j=1}^{m_{k-1}} x_{k-1,j} \big)
\end{equation}
where $d_k$ approaches zero when consecutive images are nearly identical. Further details on why this metric is designed this way, rather than using a position-wise measure, can be found in Appendix~\ref{app:metric}. To quantify the variation over the entire sequence, we average $d_k$ across all consecutive image pairs to obtain the inter-image variation metric, which is defined as follows:
\begin{equation}
\mathcal{D}_{inter} = \frac{1}{n-1} \sum_{k=2}^{n} d_k
\end{equation}
A higher $\mathcal{D}_{inter}$ indicates greater differences between images, suggesting a higher token retention ratio during inter-image pruning to preserve semantic content across images.

To determine how many visual tokens should be preserved at the intra-image and inter-image pruning stages, we estimate the relative amount of redundancy within and across images. Using the intra-image diversity $\mathcal{D}_{{intra}}$ and inter-image variation $\mathcal{D}_{{inter}}$, we introduce a factor $s$ that captures both within-image complexity and cross-image differences,  which can be represented as follows:
\begin{equation}
\label{equation:s}
s = \frac{\mathcal{D}_{{intra}}}{\mathcal{D}_{{inter}}}
\end{equation}
This factor provides an intuitive signal that when intra-image differences are large and cross-image variation is small, $s$ becomes larger, resulting in a higher retention ratio at the intra-image pruning stage. Conversely, when cross-image variation dominates, $s$ decreases, indicating that fewer tokens should be kept within each image so that more capacity can be allocated to modeling global semantic changes across images. We map this factor to an actual retention size using a transformation. Given the minimum and maximum intra-image budgets, $M_{min}$ and $M_{max}$, and a scaling coefficient $\lambda$ controlling the influence of $s$, we can define the target retention size \(M_1\) for the intra-image pruning stage as follows:
\begin{equation}
M_1 = M_{min} + \lfloor (M_{max} - M_{min}) \cdot \lambda \cdot s \rceil
\end{equation}
Notably, we clip $\lambda \cdot s$ to the range [0, 1]. More details on the distribution of the $s$ values can be found in Appendix~\ref{app:implement}. This design allows adaptive retention. Experiments show our method performs well across various datasets and parameter settings, demonstrating strong generalization.

\subsection{Intra-Image Processing Stage}
The intra-image pruning stage aims to remove redundant tokens within each image, retaining tokens that best capture the core visual structure. We assign a {retention weight} to each image based on its intra-image diversity. For image $I_k$, the weight \(w_k\) is defined as follows:
\begin{equation}
w_k = \mathcal{D}_{{intra}}(I_k)
\end{equation}
where a higher weight indicates that the image requires more tokens to be retained. The last image in a sequence often carries more importance for the response and receives higher attention. Thus, when there are more than two images (for two images, which often correspond to each other, their relative importance is typically comparable), the retention weight of the last image is set as follows:
\begin{equation}
\label{equation:last}
w_n = \max_k w_k \quad \text{if } n > 2
\end{equation}
Given these weights, the number of tokens \(m_k^{(1)}\) to retain for image $I_k$ during intra-image pruning is computed as follows:
\begin{equation}
m_k^{(1)} =\lfloor \frac{w_k}{\sum_{j=1}^{n} w_j} \cdot {M_1} \rceil
\end{equation}
With the retention count determined for each image, we apply \(GreedyRepMax\) (Equation \ref{div}) to select the $m_k^{(1)}$ most representative tokens \(X_k^{(1)}\) from the original token set $X_k$, as follows:
\begin{equation}
X_k^{(1)} = {GreedyRepMax}(X_k, m_k^{(1)})
\end{equation}
The output \(X^{(1)}\) of this stage can be expressed as follows:
\begin{equation}
X^{(1)} = \bigcup_{k=1}^{n} X_k^{(1)}, \; \, |X^{(1)}| = M_1
\end{equation}

\subsection{Inter-Image Processing Stage}
The inter-image pruning stage aims to extract the most critical semantic information across all images, obtaining the final set \(X^{(*)}\) of $M_{{final}}$ tokens. We first apply a global \(GreedyRepMax\) on $X^{(1)}$ to maximize the coverage of token representations and retain $M_2$ tokens, with result \(X^{(2)}\) as follows:
\begin{equation}
X^{(2)} = {GreedyRepMax}(X^{(1)}, M_2)
\end{equation}
Next, to further select tokens that contribute most across images and are well-aligned with the text, we use two metrics for each token \(x_i\), namely inter-image diversity \(v_i\) and text alignment \(a_i\). These metrics are defined as follows:
\begin{equation}
v_i = \frac{1}{|X^{(2)}|-1} \sum_{j \neq i} (1 - \cos(x_i, x_j))
\end{equation}
\begin{equation}
a_i = -\frac{1}{|X^t|} \sum_{t \in X^t} \lVert x_i - t \rVert^2_2
\end{equation}
Here, $X^t$ is the textual token set. 
We use \(Pareto\) selection to choose the set of tokens from $X^{(2)}$ that are optimal in both metrics. More details on \(Pareto\) selection can be found in Appendix~\ref{appendix:Pareto}. The result \(X^{(*)}\) can be represented as follows:
\begin{equation}
X^{(*)} = {Pareto}(X^{(2)}, \{v_i\}, \{a_i\}, M_{{final}})
\end{equation}
where $\{v_i\}$ and $\{a_i\}$ represent the inter-image diversity and text alignment scores of all tokens in $X^{(2)}$, respectively. The resulting set $X^{(*)}$, containing $M_{{final}}$ tokens, is subsequently fed into the language model.

\definecolor{best}{RGB}{255,215,0}      
\definecolor{second}{RGB}{220,220,220}  
\definecolor{rowlight}{RGB}{248,248,248}

\newcommand{\best}[1]{\cellcolor{best!45}\textbf{#1}}

\begin{table*}[!t]
\centering
\renewcommand{\arraystretch}{1.18}
\setlength{\tabcolsep}{6pt}

\resizebox{\textwidth}{!}{
\begin{tabular}{>{\bfseries\itshape}c *{9}{c}}
\toprule
\textbf{\textit{Method}} &
\textbf{\textit{ActPred}} & \textbf{\textit{ALFRED}} &
\textbf{\textit{CFInfer}} & \textbf{\textit{IEdit}} &
\textbf{\textit{MoveDir}} & \textbf{\textit{MMQA}} &
\textbf{\textit{ObjExist}} & \textbf{\textit{ObjShuf}} &
\textbf{\textit{OCR-VQA}} \\
\midrule
\multicolumn{10}{c}{\textit{\textbf{LLaVA-1.5-7B}}} \\
\midrule
Original Model & 52.0 & 14.81 & 30.5 & 5.36 & 32.0 & 67.0 & 50.5 & 34.5 & 8.0 \\
\cdashline{1-10}
FastV          & 52.5 & 9.82  & 22.5 & 4.26 & 18.0 & 61.0 & 40.5 & 32.0 & 3.5 \\
VTW            & 52.0 & 10.28 & 22.0 & 4.37 & 19.5 & 60.5 & 42.0 & 31.5 & 2.5 \\
SparseVLM      & 53.5 & 11.06 & 23.0 & 5.03 & 20.5 & 62.0 & 44.5 & 33.0 & 3.0 \\
VisionZip      & 53.5 & 11.18 & 22.5 & 4.43 & 22.5 & 61.0 & 42.5 & 32.5 & 3.0 \\
CATP           & 51.5 & 10.32 & 21.0 & 4.21 & 17.0 & 60.5 & 39.0 & 31.5 & 3.5 \\
DivPrune       & 54.0 & 11.31 & 22.5 & 4.61 & 17.5 & 62.0 & 36.0 & 32.5 & 3.0 \\
TrimTokenator  & 54.0 & 11.26 & 23.5 & 4.52 & 22.0 & 61.5 & 43.0 & 33.0 & 3.5 \\
\rowcolor{rowlight}
\best{Ours}  &
\best{56.0} & \best{12.74} & \best{25.0} & \best{5.15} &
\best{25.5} & \best{62.5} & \best{47.0} & \best{33.5} & \best{4.5} \\
\midrule

\multicolumn{10}{c}{\textit{\textbf{LLaVA-1.5-13B}}} \\
\midrule
Original Model & 45.0 & 18.64 & 38.0 & 9.00 & 37.0 & 70.5 & 47.0 & 40.5 & 46.0 \\
\cdashline{1-10}
FastV          & 46.0 & 15.17 & 34.0 & 7.48 & 32.5 & 66.5 & 46.0 & 40.5 & 45.5 \\
VTW            & 46.5 & 15.35 & 34.5 & 7.61 & 33.5 & 66.0 & 45.5 & 39.5 & 46.0 \\
SparseVLM      & 47.0 & 16.21 & 37.0 & 7.96 & 33.5 & 67.0 & 48.0 & 41.0 & 47.5 \\
VisionZip      & 46.5 & 15.76 & 36.5 & 7.82 & 32.5 & 67.5 & 47.5 & 39.5 & 47.0 \\
CATP           & 44.5 & 15.62 & 34.0 & 7.57 & 32.0 & 66.5 & 45.5 & 40.0 & 45.5 \\
DivPrune       & 47.5 & 15.50 & 37.0 & 8.30 & 34.0 & 66.0 & 47.5 & 40.5 & 46.5 \\
TrimTokenator  & 47.0 & 16.25 & 36.5 & 8.15 & 33.5 & 67.0 & 47.0 & 39.5 & 47.0 \\
\rowcolor{rowlight}
\best{Ours}  &
\best{48.5} & \best{18.14} & \best{38.0} & \best{8.35} &
\best{35.5} & \best{69.0} & \best{51.5} & \best{42.5} & \best{49.5} \\
\midrule

\multicolumn{10}{c}{\textit{\textbf{Yi-VL-6B}}} \\
\midrule
Original Model & 40.5 & 16.02 & 36.5 & 7.90 & 21.0 & 67.0 & 53.0 & 36.5 & 27.5 \\
\cdashline{1-10}
FastV          & 39.0 & 17.62 & 31.0 & 7.12 & 28.5 & 63.0 & 50.5 & 30.5 & 23.0 \\
VTW            & 40.0 & 17.87 & 31.5 & 7.25 & 27.0 & 65.5 & 51.0 & 30.0 & 23.5 \\
SparseVLM      & 39.5 & 18.35 & 32.0 & 7.18 & 29.0 & 64.0 & 52.0 & 30.5 & 24.5 \\
VisionZip      & 40.5 & 18.17 & 32.5 & 7.29 & 28.0 & 63.0 & 53.5 & 31.0 & 24.0 \\
CATP           & 39.5 & 17.82 & 31.5 & 7.15 & 28.5 & 63.5 & 51.5 & 29.5 & 23.5 \\
DivPrune       & 40.0 & 18.76 & 33.0 & 7.45 & 29.5 & 66.0 & 55.0 & 31.5 & 25.0 \\
TrimTokenator  & 39.5 & 18.67 & 32.5 & 7.40 & 29.0 & 64.5 & 53.5 & 30.5 & 24.0 \\
\rowcolor{rowlight}
\best{Ours}  &
\best{40.5} & \best{19.84} & \best{34.0} & \best{7.59} &
\best{30.0} & \best{67.5} & \best{55.5} & \best{32.5} & \best{25.5} \\
\bottomrule
\end{tabular}}
\caption{Performance comparison of different methods across multiple models and tasks at a retention ratio of 0.2.}
\label{tab:main_result}
\end{table*}

\section{Experiments}
\subsection{Experimental Setup}
\textbf{Models and Baselines. }To evaluate the effectiveness of our method, we perform extensive experiments on a diverse set of representative LMMs, including LLaVA-1.5-7B \cite{liu2023llava1.5}, LLaVA-1.5-13B \cite{liu2023llava1.5}, Yi-VL-6B \cite{young2024yi-VL} and InternVL-Chat-ViT-6B-Vicuna-13B \cite{chen2024expanding}. These models span a wide range of parameter scales and architectures. We compare our method against a set of widely used pruning baselines, including FastV \cite{chen2024fastv}, VTW \cite{lin2025boosting}, SparseVLM \cite{zhang2024sparsevlm}, VisionZip \cite{yang2025visionzip}, CATP \cite{li2025catp}, DivPrune \cite{alvar2025divprune} and TrimTokenator \cite{zhang2025trimtokenator}. All baselines are assessed under consistent experimental settings to ensure a rigorous and fair comparison.

\textbf{Datasets and Metrics. }We evaluate our method on a series of benchmark datasets, including ActionPrediction (ActPred) \cite{wu2024star}, ALFRED \cite{shridhar2020alfred}, CounterfactualInference (CFInfer) \cite{yi2019clevrer}, IEdit \cite{tan2019expressing}, MovingDirection (MoveDir) \cite{yi2019clevrer}, MultiModalQA (MMQA) \cite{talmor2021multimodalqa}, ObjectExistence (ObjExist) \cite{yi2019clevrer}, ObjectShuffle (ObjShuf) \cite{patraucean2023perception} and OCR-VQA \cite{mishra2019ocr}. These datasets include from just a few to over a hundred images and cover diverse tasks, offering a comprehensive evaluation setting. We employ accuracy to evaluate performance on understanding datasets, and Rouge-L for generative datasets.



\textbf{Implementation Details. }{We utilize an NVIDIA H100 GPU with 80GB of memory.} We set the default values of the parameters \(M_{min}\), \(M_{max}\), \(\lambda\) and \(M_2\) to 294, 454, 0.5 and 252, respectively. We set the default token retention ratio to 0.2 in both the comparative and ablation experiments. More details can be found in Appendix \ref{app:implement}.


\subsection{Main Results}

Table \ref{tab:main_result} reports the comparative performance of our method across multiple LMMs under a token retention ratio of 0.2. We evaluate it against the original model as well as a range of token pruning methods. The results demonstrate that our method consistently outperforms competing approaches across different model scales and tasks, highlighting its effectiveness in long context multimodal scenarios. On LLaVA-1.5-7B, our method consistently surpasses strong baselines such as FastV, VTW and DivPrune. Specifically, for the ActionPrediction task, our method achieves a performance of 56.0, showing a clear improvement over FastV and TrimTokenator. Moreover, in some cases, performance even exceeds the original model, likely due to denser critical information and less irrelevant visual noise. Our approach exhibits remarkable robustness: on LLaVA-1.5-13B, it outperforms the original model across tasks such as ObjectShuffle and OCR-VQA. Furthermore, our method demonstrates strong generalization capabilities, achieving excellent performance on the Yi-VL-6B architecture as well. At the same token retention ratio, it consistently surpasses other baselines across a wide range of tasks. These findings validate the robustness of our method across different model architectures and its scalability to models of various sizes. Additionally, we provide the performance comparison of InternVL-Chat-ViT-6B-Vicuna-13B in Appendix \ref{Vicuna}. We also provide the performance comparison of various methods under a different token retention ratio in Appendix \ref{app:0.5}.

\subsection{Efficiency Analysis (Accounting for Pruning Overhead)}
We conduct efficiency tests on LLaVA-1.5-7B in FP16 precision using images with lengths ranging from 2 to 32, with inputs also in FP16, and a batch size of 1. Specifically, under the same pruning ratio as in the main experiments, we measure inference latency and GPU memory usage. It is important to emphasize that the reported latency corresponds to the prefill stage, as the pruning operations are performed during this stage. By the time the decoding stage begins, all pruning is already complete; therefore, the measured latency fully accounts for the overhead introduced by the pruning process. As presented in Table \ref{tab:efficiency}, our method achieves acceleration in inference and a notable reduction in memory consumption, offering a practical and effective solution for deploying long context multimodal models in resource-constrained environments. More details on pruning acceleration can be found in Appendix~\ref{app:efficiency}.
\begin{table*}[!t]
\centering
\resizebox{\textwidth}{!}{%
\begin{tabular}{c|cccccccc}
\toprule
\textbf{{Method}} & \textbf{\textit{Original}} & \textbf{\textit{FastV}} & \textbf{\textit{VTW}} & \textbf{\textit{SparseVLM}} & \textbf{\textit{VisionZip}} & \textbf{\textit{CATP}} & \textbf{\textit{DivPrune}} & \textbf{\textit{Ours}} \\ \midrule
\textbf{{Latency (ms)}} & 1035.47 & 825.91 & 813.85 & 821.01 & 813.05 & 827.92 & 810.46 & 802.65 \\
\textbf{{Memory (G)}} & 19.38 & 15.34 & 15.27 & 15.33 & 15.20 & 15.37 & 15.20 & 15.18 \\ \bottomrule
\end{tabular}%
}
\caption{Comparison of GPU memory and latency for our pruning method and baselines on LLaVA-1.5-7B.}
\label{tab:efficiency}
\end{table*}

\subsection{Ablation Study}
To comprehensively analyze the contribution of each component in our method, we conduct extensive ablation studies on LLaVA-1.5-7B, with the results summarized in Table~\ref{tab:ablation_study}. We first examine the necessity of the two stage architecture and its sequential design. Specifically, we consider three ablation settings: \(Ablation_1\), which completely removes the intra-image pruning stage; \(Ablation_2\), which removes the inter-image pruning stage; and \(Ablation_3\), which swaps the execution order by performing inter-image pruning before intra-image. As shown in the results, all three variants lead to consistent performance degradation. These findings indicate that eliminating local redundancy is a prerequisite for effective global modeling, and that modeling cross-image variations is crucial for long context understanding. Moreover, token pruning must first be conducted at the single image level to obtain compact image representations, upon which accurate global pruning of cross-image redundancy can be performed. We further investigate the impact of the dynamic budget allocation and the text alignment pareto selection strategy. In \(Ablation_4\), we remove dynamic budget allocation and instead assign a fixed and uniform budget to each image. In \(Ablation_5\), we disable the pareto selection mechanism entirely. In \(Ablation_6\), pareto selection is prematurely applied during the intra-image pruning stage while being omitted in the inter-image stage. All these ablations result in performance drops, demonstrating the importance of dynamic budget allocation as well as applying pareto selection at the inter-image level. More ablation results can be found in Appendix~\ref{app:ablation_results}.

\begin{table}[h]
\centering
\resizebox{0.48\textwidth}{!}{%
\begin{tabular}{c|ccccc}
\toprule
\textbf{\textit{Setting}} &
\textbf{\textit{ActPred}} &
\textbf{\textit{CFInfer}} &
\textbf{\textit{MoveDir}} &
\textbf{\textit{ObjExist}} &
\textbf{\textit{OCR-VQA}} \\ \midrule
\(Ablation_1\) & 55.0 & 24.0 & 24.5 & 45.0 & 2.5 \\
\(Ablation_2\) & 54.0 & 24.0 & 25.0 & 45.5 & 4.0 \\
\(Ablation_3\) & 53.0 & 22.0 & 23.0 & 45.5 & 2.5 \\
\(Ablation_4\) & 54.5 & 23.5 & 23.0 & 45.5 & 4.0 \\
\(Ablation_5\) & 53.0 & 22.0 & 18.5 & 41.0 & 2.5 \\
\(Ablation_6\) & 55.0 & 22.5 & 19.0 & 41.0 & 4.0 \\
\textbf{Ours} & \textbf{56.0} & \textbf{25.0} & \textbf{25.5} & \textbf{47.0} & \textbf{4.5} \\ \bottomrule
\end{tabular}
}
\caption{Performance comparison under different ablation settings on LLaVA-1.5-7B across benchmarks.}
\label{tab:ablation_study}
\end{table}

\subsection{Performance on Different Retention Ratios}
\begin{figure}
    \centering
    \begin{subfigure}[t]{0.48\linewidth}
        \centering
        \includegraphics[width=\linewidth]{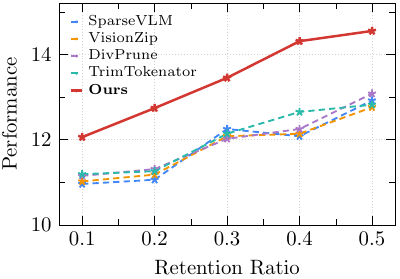}
        \caption{ALFRED}
    \end{subfigure}
    \hfill
    \begin{subfigure}[t]{0.48\linewidth}
        \centering
        \includegraphics[width=\linewidth]{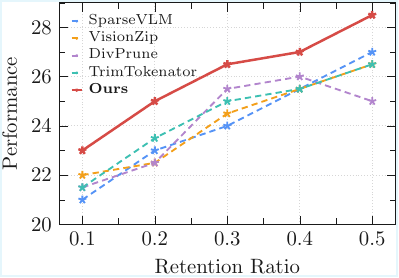}
        \caption{CFInfer}
    \end{subfigure}
    \caption{Performance across different token retention ratios with LLaVA-1.5-7B.}
    \label{fig:Token_retention_ration}
\end{figure}

We analyze the performance trends of LLaVA-1.5-7B on the representative tasks ALFRED and CFInfer under different visual token retention ratios, ranging from 0.1 to 0.5. As shown in Figure~\ref{fig:Token_retention_ration}, our method consistently and significantly outperforms all baseline methods across all retention settings. Specifically, even under the most stringent retention ratio of 0.1, our method achieves a performance of 12.06 on ALFRED, substantially surpassing the others. This notable gap indicates that our approach effectively preserves crucial information, whereas strategies that rely solely on local information tend to struggle with this task. Furthermore, as the retention ratio increases to 0.5, our performance advantage is not only maintained but further widened, clearly exceeding other methods and demonstrating the robustness of our approach.


\section{Conclusion}

In this paper, we explore the challenges of visual token pruning in long context multi-image inputs. Based on this, we propose an adaptive pruning method that uses intra-image diversity and inter-image variation to model the two types of redundancy. Our method consists of two stages: intra-image pruning and inter-image pruning. The intra-image pruning allocates token budgets adaptively based on each image’s visual richness and uses a greedy strategy to select the most representative tokens for each image. The inter-image pruning performs global diversity filtering over all image tokens to form a candidate set and then applies Pareto selection based on diversity and text alignment to retain the most informative and text-relevant tokens. Extensive experiments demonstrate the soundness and effectiveness of our method.

\section*{Limitations}
In this work, we conduct experiments across multiple models and datasets to evaluate the effectiveness of our proposed multimodal long context visual token pruning method. The results demonstrate the validity of our approach. However, due to computational constraints, we have not yet evaluated it on extremely large models, such as those with 70 billion parameters. Exploring the scalability of our method to such large models constitutes an important direction for future work.


\bibliography{custom}
\clearpage
\appendix

\section{Algorithm Overview}
We outline the overall procedure of TrimTokenator-LC in Algorithm~\ref{alg:adaptive_pruning}, which provides the pseudocode of our proposed method.

\begin{algorithm}[!b]
\caption{\textbf{TrimTokenator-LC}}
\label{alg:adaptive_pruning}
\begin{algorithmic}[1]

\REQUIRE Visual token sets $(X_1,\dots,X_n)$, where $X_k=\{x_{k,1},\dots,x_{k,m_k}\}$;
text tokens $X_t$; final budget $M_{{final}}$;
budgets $M_{min}$ and $M_{max}$, scaling coefficient $\lambda$ and candidate budget $M_2$
\ENSURE Final visual token set $X^{(*)}$

\vspace{0.5em}
\STATE \textbf{/* Compute Redundancy Signals */}
\FOR{$k=1$ \TO $n$}
    \STATE Compute Intra-Image Diversity $\mathcal{D}_{intra}(I_k)$
\ENDFOR
\STATE $\mathcal{D}_{intra} \leftarrow \frac{1}{n}\sum_{k=1}^n \mathcal{D}_{intra}(I_k)$
\FOR{$k=2$ \TO $n$}
    \STATE Compute Inter-Image Variation $d_k$
\ENDFOR
\STATE $\mathcal{D}_{inter} \leftarrow \frac{1}{n-1}\sum_{k=2}^n d_k$
\STATE $s \leftarrow \mathcal{D}_{intra} / \mathcal{D}_{inter}$
\STATE $M_1 \leftarrow M_{min} + \lfloor (M_{max}-M_{min})\cdot \lambda \cdot s \rceil$

\vspace{0.5em}
\STATE \textbf{/* Intra-Image Processing Stage */}
\FOR{$k=1$ \TO $n$}
    \STATE $w_k \leftarrow \mathcal{D}_{intra}(I_k)$
\ENDFOR
\IF{$n > 2$}
    \STATE $w_n \leftarrow \max_k w_k$
\ENDIF
\FOR{$k=1$ \TO $n$}
    \STATE $m_k^{(1)} \leftarrow \lfloor \frac{w_k}{\sum_{j=1}^{n} w_j} \cdot M_1 \rceil$
    \STATE $X_k^{(1)} \leftarrow \textit{GreedyRepMax}(X_k, m_k^{(1)})$
\ENDFOR
\STATE $X^{(1)} \leftarrow \bigcup_{k=1}^n X_k^{(1)}$

\vspace{0.5em}
\STATE \textbf{/* Inter-Image Processing Stage */}
\STATE $X^{(2)} \leftarrow \textit{GreedyRepMax}(X^{(1)}, M_2)$
\FORALL{$x_i \in X^{(2)}$}
    \STATE $v_i \leftarrow \frac{1}{|X^{(2)}|-1}\sum_{j\neq i}\left(1-\cos(x_i,x_j)\right)$
    \STATE $a_i \leftarrow -\frac{1}{|X^t|}\sum_{t\in X^t}\|x_i-t\|_2^2$
\ENDFOR
\STATE $X^{(*)} \leftarrow \textit{Pareto}(X^{(2)}, \{v_i\}, \{a_i\}, M_{{final}})$

\RETURN $X^{(*)}$

\end{algorithmic}
\end{algorithm}

\section{Details of Pareto Selection}
\label{appendix:Pareto}

{Pareto selection} is a multi-objective optimization method used to select an optimal set of solutions when multiple conflicting objectives exist. Its core idea is based on the concept of a {non-dominated solution}. A solution is considered non-dominated if no other solution is better in all objectives simultaneously. All non-dominated solutions form the {Pareto front}, representing the trade-off between different objectives. Formally, given two objective functions $f_1(x)$ and $f_2(x)$, a candidate solution $x_i$ is Pareto optimal if there does not exist another solution $x_j$ such that $f_1(x_j) \ge f_1(x_i)$ and $f_2(x_j) \ge f_2(x_i)$, with at least one strict inequality, that is, $(f_1(x_j) > f_1(x_i) \text{ or } f_2(x_j) > f_2(x_i))$. More details on how this problem is addressed can be found in Appendix~\ref{app:efficiency}.

\begin{table*}[!t]
\centering
\renewcommand{\arraystretch}{1.18}
\setlength{\tabcolsep}{6pt}

\resizebox{\textwidth}{!}{
\begin{tabular}{>{\bfseries\itshape}c *{9}{c}}
\toprule
\textbf{\textit{Method}} &
\textbf{\textit{ActPred}} &
\textbf{\textit{ALFRED}} &
\textbf{\textit{CFInfer}} &
\textbf{\textit{IEdit}} &
\textbf{\textit{MoveDir}} &
\textbf{\textit{MMQA}} &
\textbf{\textit{ObjExist}} &
\textbf{\textit{ObjShuf}} &
\textbf{\textit{OCR-VQA}} \\
\midrule
Original Model & 51.0 & 16.20 & 30.0 & 10.74 & 29.0 & 66.0 & 50.5 & 37.0 & 16.5 \\
\cdashline{1-10}
FastV          & 48.0 & 13.73 & 28.5 & 9.82 & 33.0 & 58.0 & 45.5 & 36.0 & 10.5 \\
VTW            & 49.0 & 13.62 & 27.5 & 9.75 & 32.5 & 59.5 & 46.0 & 36.5 & 11.5 \\
SparseVLM      & 50.5 & 14.94 & 28.0 & 10.32 & 32.0 & 60.5 & 46.5 & 37.0 & 11.0 \\
VisionZip      & 52.0 & 14.76 & 28.5 & 10.28 & 33.5 & 58.5 & 45.5 & 35.5 & 11.5 \\
CATP           & 51.0 & 14.15 & 25.0 & 10.12 & 32.5 & 58.0 & 45.0 & 35.5 & 12.0 \\
DivPrune       & 52.0 & 15.23 & 26.5 & 10.77 & 34.0 & 62.5 & 48.5 & 37.0 & 12.5 \\
TrimTokenator  & 51.5 & 15.12 & 26.0 & 10.61 & 32.0 & 62.0 & 48.0 & 36.0 & 11.5 \\
\rowcolor{rowlight}
\best{Ours} &
\best{54.0} & \best{16.10} & \best{30.0} & \best{11.34} &
\best{34.5} & \best{63.0} & \best{49.0} & \best{37.5} & \best{13.0} \\
\bottomrule
\end{tabular}}
\caption{Performance comparison of visual token pruning methods on InternVL-Chat-ViT-6B-Vicuna-13B.}
\label{tab:more_result}
\end{table*}

\section{Metric Analysis}
\label{app:metric}
Intra-image diversity is designed to quantify the degree of dispersion among visual tokens within a single image in the embedding space. This metric is motivated by the need for minimal semantic coverage per image in long context settings. During visual encoding, an image is typically mapped to a large number of local tokens, many of which originate from semantically or structurally similar regions whose marginal information contribution quickly diminishes. When an image is dominated by large homogeneous backgrounds or repetitive textures, token representations tend to be highly similar, resulting in low intra-image diversity and indicating substantial internal redundancy. In contrast, images containing multiple objects or complex spatial structures yield more dispersed token representations in the embedding space, leading to higher intra-image diversity. Defined as the average cosine distance among tokens within the same image, this metric directly reflects the number of independent visual prototypes present in the image. It therefore provides a quantitative criterion for determining whether more tokens should be preserved during pruning to maintain the image’s fundamental semantic structure, effectively preventing excessive compression of visually complex images.


Inter-image variation is designed to characterize the degree of semantic change between images in long context inputs. Treating all image tokens as independent is unreasonable when the visual content is redundant. To address this issue, we quantify the amount of new information introduced by the current image $I_k$ relative to its predecessor $I_{k-1}$ by comparing their global semantic embeddings. Specifically, we aggregate all visual tokens of each image into a single global representation and compute the cosine distance between consecutive images. When neighboring images are highly similar or repetitive, their global embeddings are strongly aligned, yielding a small inter-image variation value, which indicates that cross-image redundancy dominates. Because this metric captures global semantic transitions across images rather than local token fluctuations, it provides a stable and direct quantitative criterion for preserving the most critical visual information needed to model semantic evolution in long context scenarios. Furthermore, we consider methods that estimate inter-image changes based on position-wise differences of visual tokens, which can represent as follows:
\begin{equation}
d_k = \frac{1}{m_k} \sum_{i=1}^{m_k} \big(1 - \cos(x_{k,i}, x_{k-1,i})\big)
\end{equation}
Such approaches implicitly assume that the spatial structure across different images is consistently aligned, i.e., tokens at the same index are semantically comparable. However, this assumption often does not hold in long context scenarios: images from different sources (e.g., webpages, slides or unrelated photos) generally lack strict spatial correspondence; even consecutive images from the same source may experience spatial shifts due to viewpoint changes, content rearrangement or significant object motion, making "corresponding tokens" no longer truly semantically aligned. Relying on position-wise matching in such cases can not only introduce noisy estimates but also systematically underestimate the true cross-image semantic changes. Moreover, in some tasks, consecutive images may be entirely different (e.g., two different documents or unrelated objects). Comparing the top-left region of a document with that of a natural image yields a ‘variation’ score driven by spatial misalignment rather than semantic redundancy.

\section{Performance Comparison on InternVL-Chat-ViT-6B-Vicuna-13B}
\label{Vicuna}
To further validate the generality of our method, we extend our evaluation to the InternVL-Chat-ViT-6B-Vicuna-13B model, which employs a substantially larger visual encoder than the LLaVA family. As shown in Table \ref{tab:more_result}, our method consistently outperforms the baselines on this architecture, achieving the best performance on various tasks. By selectively retaining visually informative tokens, our approach effectively reduces computational overhead while improving the quality of the preserved visual context. These results further demonstrate the strong robustness of our method and highlight its ability to generalize effectively across different model architectures.

\section{Performance Comparison under a Different Token Retention Ratio}
\label{app:0.5}
We further evaluate the model performance on LLaVA-1.5-7B under a higher token retention ratio of 0.5. As shown in Table~\ref{tab:0.5_prune_ration}, our method continues to outperform all other pruning strategies, demonstrating that it maintains a clear advantage even under relatively high token retention settings. This result highlights the robustness of our approach, indicating that selectively preserving informative tokens can consistently enhance performance without being overly sensitive to the retention ratio.

\begin{table*}[!t]
\centering
\renewcommand{\arraystretch}{1.18}
\setlength{\tabcolsep}{6pt}

\resizebox{\textwidth}{!}{
\begin{tabular}{>{\bfseries\itshape}c *{9}{c}}
\toprule
\textbf{\textit{Method}} &
\textbf{\textit{ActPred}} &
\textbf{\textit{ALFRED}} &
\textbf{\textit{CFInfer}} &
\textbf{\textit{IEdit}} &
\textbf{\textit{MoveDir}} &
\textbf{\textit{MMQA}} &
\textbf{\textit{ObjExist}} &
\textbf{\textit{ObjShuf}} &
\textbf{\textit{OCR-VQA}} \\
\midrule
Original Model & 52.0 & 14.81 & 30.5 & 5.36 & 32.0 & 67.0 & 50.5 & 34.5 & 8.0 \\
\cdashline{1-10}
FastV          & 46.0 & 11.68 & 25.5 & 6.85 & 25.0 & 63.0 & 46.5 & 16.0 & 6.0 \\
VTW            & 47.5 & 11.92 & 25.0 & 7.03 & 25.5 & 64.5 & 46.0 & 15.5 & 7.0 \\
SparseVLM      & 49.0 & 12.92 & 27.0 & 7.18 & 26.0 & 65.5 & 47.5 & 16.5 & 6.5 \\
VisionZip      & 48.5 & 12.76 & 26.5 & 7.24 & 26.5 & 65.5 & 48.5 & 16.5 & 7.5 \\
CATP           & 47.5 & 12.07 & 25.5 & 7.17 & 25.5 & 63.0 & 46.5 & 16.0 & 6.0 \\
DivPrune       & 48.5 & 13.08 & 25.0 & 7.35 & 27.0 & \textbf{66.5} & 48.5 & 16.5 & 7.5 \\
TrimTokenator  & 47.0 & 12.82 & 26.5 & 7.25 & 26.5 & 65.0 & 47.0 & 17.0 & 8.0 \\
\rowcolor{rowlight}
\best{Ours} &
\best{50.0} & \best{14.55} & \best{28.5} & \best{8.10} &
\best{27.5} & \cellcolor[RGB]{255,237,140}{65.5} & \best{50.0} & \best{18.5} & \best{9.0} \\
\bottomrule
\end{tabular}}
\caption{Comparative results of different pruning methods on LLaVA-1.5-7B under a token retention ratio of 0.5.}
\label{tab:0.5_prune_ration}
\end{table*}

\section{More Ablation Results}
\label{app:ablation_results}
We conduct additional ablation studies to further analyze our design choices. Specifically, during the adaptive allocation of the image budget, we apply a special treatment to the last image (Eq.~\ref{equation:last}). We define a variant that removes this component as \(Allocation_1\). In addition, we enforce an equal number of pruned tokens for intra-image and inter-image pruning, with half of the tokens pruned at each stage. This allocation strategy is denoted as \(Allocation_2\), and is used to evaluate the effectiveness of our two stage adaptive allocation scheme. As shown in Table~\ref{app:ablatipn_table}, both allocation strategies lead to performance degradation to varying degrees. On OCR-VQA, our method performs comparably to \(Allocation_1\), which is reasonable since most samples in this dataset contain only two images. These results validate the effectiveness of our design.
\begin{table}[h]
\centering
\resizebox{0.48\textwidth}{!}{%
\begin{tabular}{c|ccccc}
\toprule
\textbf{\textit{Setting}} &
\textbf{\textit{ActPred}} &
\textbf{\textit{CFInfer}} &
\textbf{\textit{MoveDir}} &
\textbf{\textit{ObjExist}} &
\textbf{\textit{OCR-VQA}} \\
\midrule
\(Allocation_1\) & 53.0 & 22.0 & 23.5 & 42.0 & \textbf{4.5} \\
\(Allocation_2\) & 54.5 & 24.0 & 23.5 & 45.0 & 3.5 \\
\textbf{Ours} & \textbf{56.0} & \textbf{25.0} & \textbf{25.5} & \textbf{47.0} & \textbf{4.5} \\
\bottomrule
\end{tabular}
}
\caption{Performance comparison across different allocation settings.}
\label{app:ablatipn_table}
\end{table}

\section{Hyperparameter Analysis} 

To evaluate the sensitivity of our method to hyperparameters, we examine three different configurations, including the first stage token budget bounds ($M_{min}, M_{max}$), the candidate budget $M_2$ and the scaling factor $\lambda$. Specifically, \(Hyper_1\) ($M_{min}=294, M_{max}=454, M_2=252, \lambda=0.5$) represents a high variance setting, where the token budget is heavily influenced by the factor $s$. In contrast, \(Hyper_3\) ($M_{min}=374, M_{max}=454, M_2=317, \lambda=0.2$) corresponds to a conservative strategy, featuring a higher minimum budget and reduced sensitivity to scaling. \(Hyper_2\) ($M_{min}=334, M_{max}=374, M_2=226, \lambda=0.3$) serves as an intermediate configuration, balancing the two extremes. We implement these three configurations on LLaVA-1.5-7B with a token retention rate of 0.3. As shown in Table~\ref{tab:parameters}, the results exhibit only minor differences across these configurations. Our method consistently achieves stable and strong performance under all settings, further demonstrating its robustness to variations in hyperparameters.

\begin{table}[htbp]
\centering
\resizebox{0.48\textwidth}{!}{%
\begin{tabular}{c|ccccc}
\toprule
\textbf{\textit{Setting}} & \textbf{\textit{ActPred}} & \textbf{\textit{CFInfer}} & \textbf{\textit{MMQA}} & \textbf{\textit{ObjExist}} & \textbf{\textit{OCR-VQA}} \\ \midrule
\(Hyper_1\) & 55.0 & 17.5 & 62.0 & 48.0 & 8.5 \\
\(Hyper_2\) & 55.0 & 17.5 & 61.0 & 49.0 & 8.5 \\
\(Hyper_3\) & 55.5 & 17.0 & 60.0 & 48.0 & 9.0 \\ \bottomrule
\end{tabular}
}
\caption{Performance on LLaVA-1.5-7B under different hyperparameter settings.}
\label{tab:parameters}
\end{table}

\section{Implementation Details}
\label{app:implement}
{Our experiments are conducted using the PyTorch framework \cite{paszke2019pytorch} and the Hugging Face Transformers library \cite{wolf2020transformers}. We utilize an NVIDIA H100 GPU with 80GB of memory.} We adopt accuracy as the evaluation metric for ActionPrediction (ActPred), CounterfactualInference (CFInfer), MovingDirection (MoveDir), MultiModalQA (MMQA), ObjectExistence (ObjExist), ObjectShuffle (ObjShuf) and OCR-VQA, while Rouge-L is used to evaluate the ALFRED and IEdit datasets. Our evaluation also includes some shorter multi-image datasets, such as OCR-VQA, where the number of images is typically 2 to 3. We evaluate our method across multiple datasets and find that the values of $s$ (Eq.~\ref{equation:s}) mostly lie in the range of 1 to 2. We set the default values of the parameters \(M_{min}\), \(M_{max}\), \(\lambda\) and \(M_2\) to 294, 454, 0.5 and 252, respectively. We also conduct a hyperparameter analysis, which shows that satisfactory performance can be achieved across different hyperparameter settings. We set the token retention ratio to 0.2 in both the comparative and ablation experiments.

\section{Pruning Analysis}
\paragraph{Analysis of Pruned Regions}We analyze the visual token pruning behavior under long context inputs and find that this method tends to prune regions that are highly redundant within a single image, semantically low-density, and repeatedly appear across multiple images, such as large background areas, repetitive textures, and local regions with minimal variation between adjacent images. These regions exhibit high similarity in the feature space and contribute little additional semantic information, and are therefore prioritized for compression. In contrast, regions that are semantically irreplaceable are more likely to be retained, including those that capture the core structural content of an image, regions exhibiting significant changes across images, and local visual cues that are strongly aligned with the textual context. Such regions are more widely dispersed in the embedding space and play a more crucial role, which in turn leads to their preferential retention in the final multi-image visual representation.

\paragraph{Performance on Challenging Samples}Our evaluation dataset contains a large number of highly challenging samples, such as inputs with vague and semantically sparse textual descriptions (e.g., “Please provide a common theme for these input images”), as well as images in which the key semantic elements occupy only a very small region. Strong performance on these benchmarks demonstrates that our method can robustly preserve critical visual information and maintain reliable reasoning capabilities even under such difficult input conditions.

\paragraph{Evaluation over Multiple Runs}Under the evaluation framework adopted in this work and the runtime environment described in the paper, the execution of the model is fully deterministic. As a result, repeated runs under the same settings do not introduce any randomness or variability in the outcomes. We conducted multiple independent runs using identical configurations, and the results were completely consistent across all trials, with no observable fluctuations in any of the reported metrics. This confirms that the reported performance is stable and reproducible under the specified experimental conditions.

\section{Acceleration Details}
\label{app:efficiency}
To compute the diversity metric $\mathcal{D}_{{intra}}(I)$, directly enumerating all token pairs incurs quadratic computational cost. We therefore reformulate it into an equivalent linear form. Let the normalized visual tokens of image $I$ be $\{x_i^{(I)}\}_{i=1}^{N}$, and define the aggregate representation as follows:
\begin{equation}
S_I = \sum_{i=1}^{N} x_i^{(I)}
\end{equation}
Using the identity, we obtain the following expressions:
\begin{equation}
\|S_I\|^2 = \langle S_I, S_I \rangle = \Big\langle \sum_{i=1}^{N} x_i^{(I)}, \sum_{j=1}^{N} x_j^{(I)} \Big\rangle
\end{equation}
\begin{equation}
\|S_I\|^2 
= \sum_{i=1}^{N} \|x_i^{(I)}\|^2 + \sum_{i \neq j} \langle x_i^{(I)}, x_j^{(I)} \rangle
\end{equation}
Due to our normalization of these vectors, we obtain the following formula:
\begin{equation}
\sum_{i=1}^{N} \|x_i^{(I)}\|^2 = \sum_{i=1}^{N} 1 = N
\end{equation}
\begin{equation}
\sum_{i \neq j} \langle x_i^{(I)}, x_j^{(I)} \rangle = \sum_{i \neq j} \cos(x_i^{(I)}, x_j^{(I)})
\end{equation}
Therefore, \(\|S_I\|^2\) can be expressed as follows:
\begin{equation}
\|S_I\|^2 = N + \sum_{i \neq j} \cos\!\left(x_i^{(I)}, x_j^{(I)}\right)
\end{equation}
The original pairwise definition is given as follows:
\begin{equation}
\small
D_{{intra}}(I) = \frac{1}{N(N-1)} \sum_{i \neq j} \bigl(1 - \cos\!\left(x_i^{(I)}, x_j^{(I)}\right)\bigr)
\end{equation}
Based on the above derivation, it can be reformulated as follows:
\begin{equation}
D_{{intra}}(I) = \frac{N(N-1) - (\|S_I\|^2 - N)}{N(N-1)}
\end{equation}
Similarly, the token diversity score \(v_i\) can be rewritten as follows:
\begin{equation}
v_i = \frac{N - (x_i^{(I)})^\top S_I}{N-1}
\end{equation}
These reformulations yield exactly the same value as the original definition, while avoiding explicit pairwise computation and reducing the complexity from $O(N^2)$ to $O(N)$. For instance, at a sequence length of 8192, it incurs only 2.88 ms of overhead and achieves a 23.65$\times$ speedup.

When computing the alignment score between visual tokens and text tokens, let $N_v$ denote the number of visual tokens, $M$ the number of text tokens, and $d$ the feature dimension of each token. A naive implementation directly computes the squared distance between each visual token and all text tokens and then averages the results, resulting in a complexity of $O(N_v \cdot M \cdot d)$. This becomes computationally expensive when either the number of visual tokens or the feature dimension is large. To improve efficiency, we exploit the algebraic expansion of the squared distance:
\begin{equation}
|x_i - t_j|^2 = |x_i|^2 + |t_j|^2 - 2 x_i^\top t_j
\end{equation}
Thus, the alignment score for each visual token $x_i$ can be rewritten as follows:
\begin{equation}
a_i = - |x_i|^2 - C + 2 x_i^\top \mu_t
\end{equation}
where $\mu_t = \frac{1}{M}\sum_j t_j$ and $C = \frac{1}{M}\sum_j |t_j|^2$ are precomputed constants. This formulation is mathematically equivalent to the naive implementation and introduces no approximation. By expanding the formula, we avoid constructing a large $[N_v, M, d]$ tensor and computing each visual-text pair individually. Instead, we only need to perform a single sum or squared sum over the text tokens and a vector dot product and norm computation for each visual token. This reduces the computational complexity from $O(N_v \cdot M \cdot d)$ to $O((N_v + M)d)$, significantly accelerating computation when the number of visual tokens or the feature dimension is large while maintaining exact numerical equivalence. For example, when $N_v, M = 8192, 128$, it achieves a 1.3$\times$ speedup.

In our pareto selection implementation, to improve computational efficiency, since only two objectives are involved, this process can be efficiently realized via a sorting and scanning strategy instead of exhaustive pairwise comparisons. Specifically, all candidate solutions are first sorted in descending order according to the first objective $f_1$. Then, we perform a single linear scan while maintaining the maximum value of the second objective $f_2$ observed so far. A candidate is selected as a pareto optimal solution if its $f_2$ value exceeds the current maximum, indicating that it is not dominated by any previously considered solution. This process yields the pareto front in $O(N \log N)$ time due to sorting, followed by an $O(N)$ scan. Moreover, this process occurs in the final step, at which point the number of tokens is already reduced and $N$ is relatively small. For example, when $N$ = 500 and the selected size is 14, it achieves a 14.76$\times$ speedup.

\section{Case Study}
\begin{figure*}[]
    \centering

    \begin{subfigure}{\linewidth}
        \centering
        \includegraphics[width=0.75\textwidth]{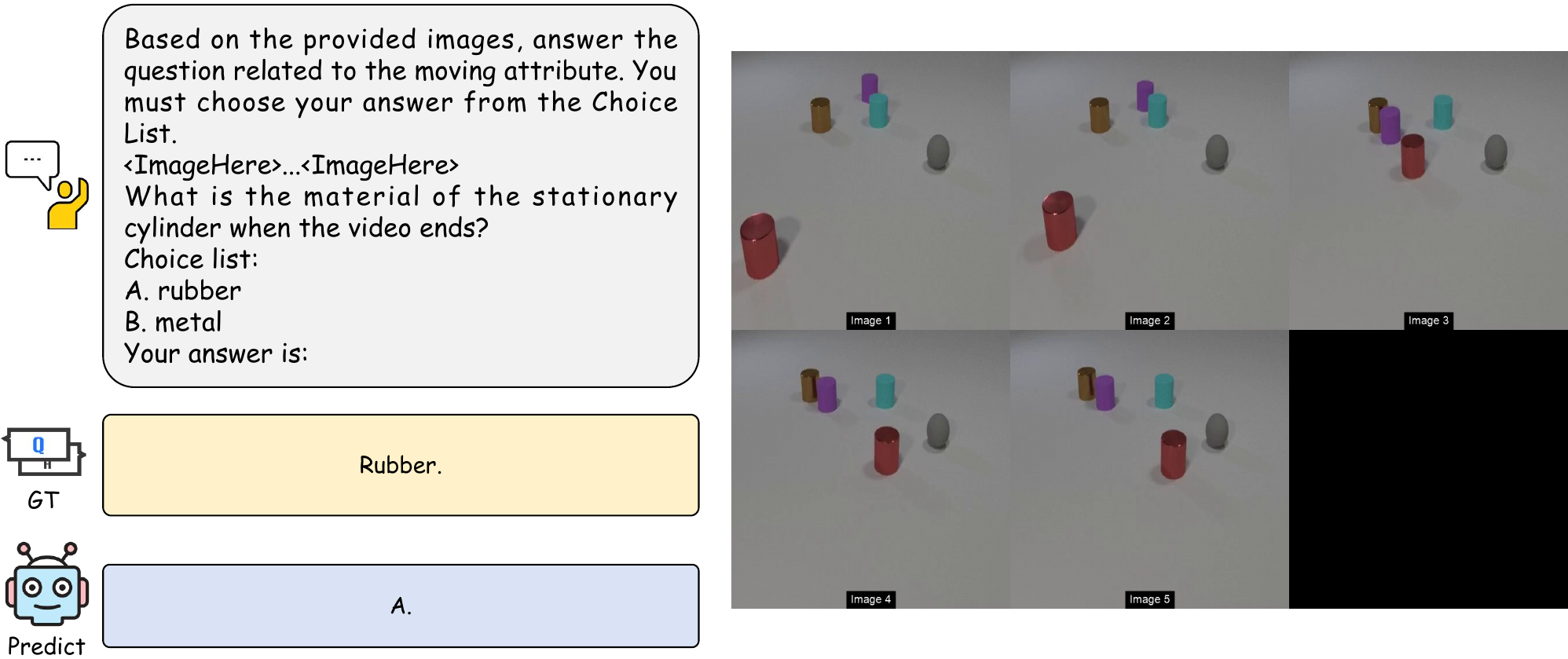}        
        \caption{}
    \end{subfigure}

    \begin{subfigure}{\linewidth}
        \centering
        \includegraphics[width=0.75\textwidth]{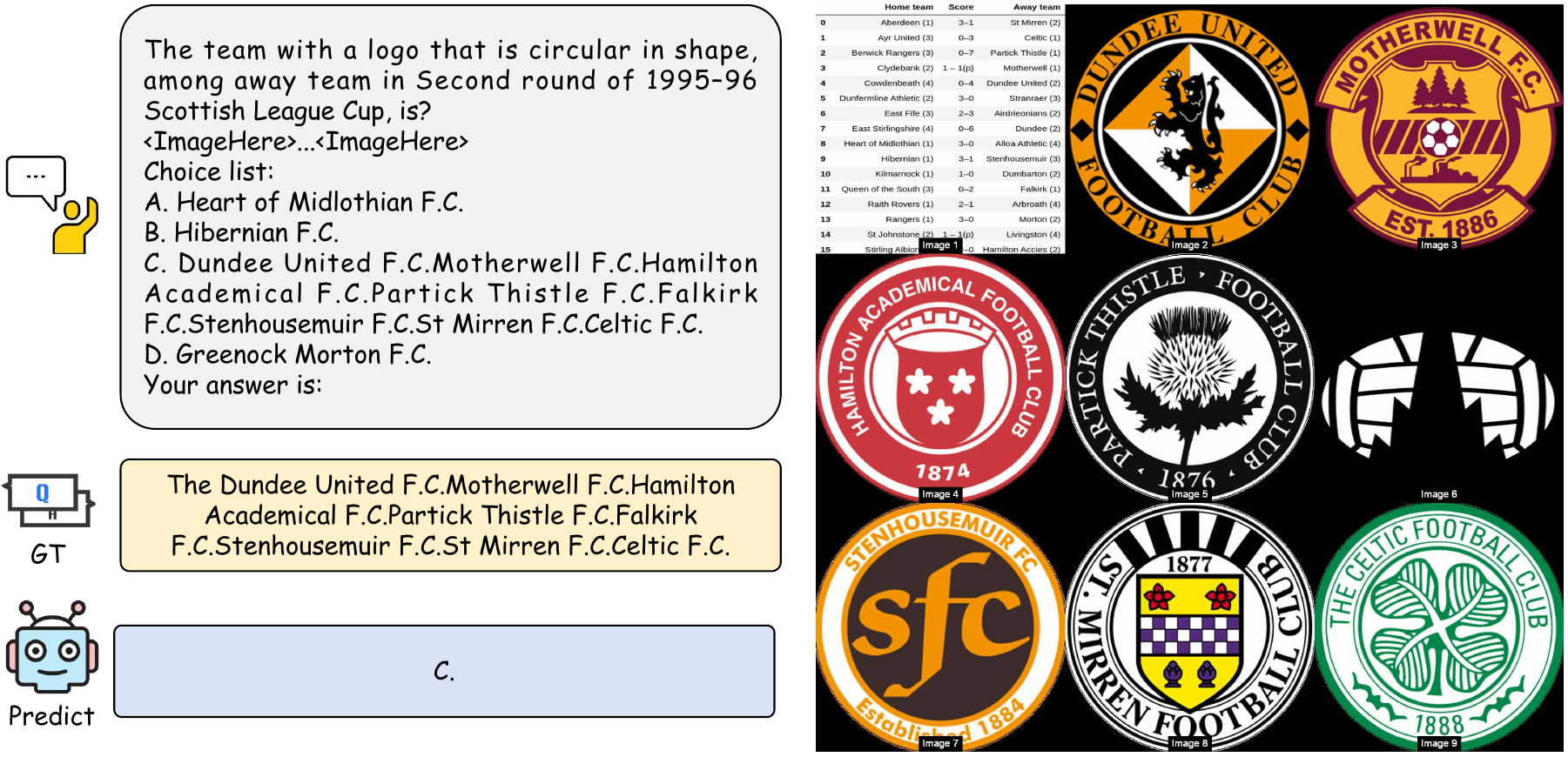}        
        \caption{}
    \end{subfigure}

    \begin{subfigure}{\linewidth}
        \centering
        \includegraphics[width=0.75\textwidth]{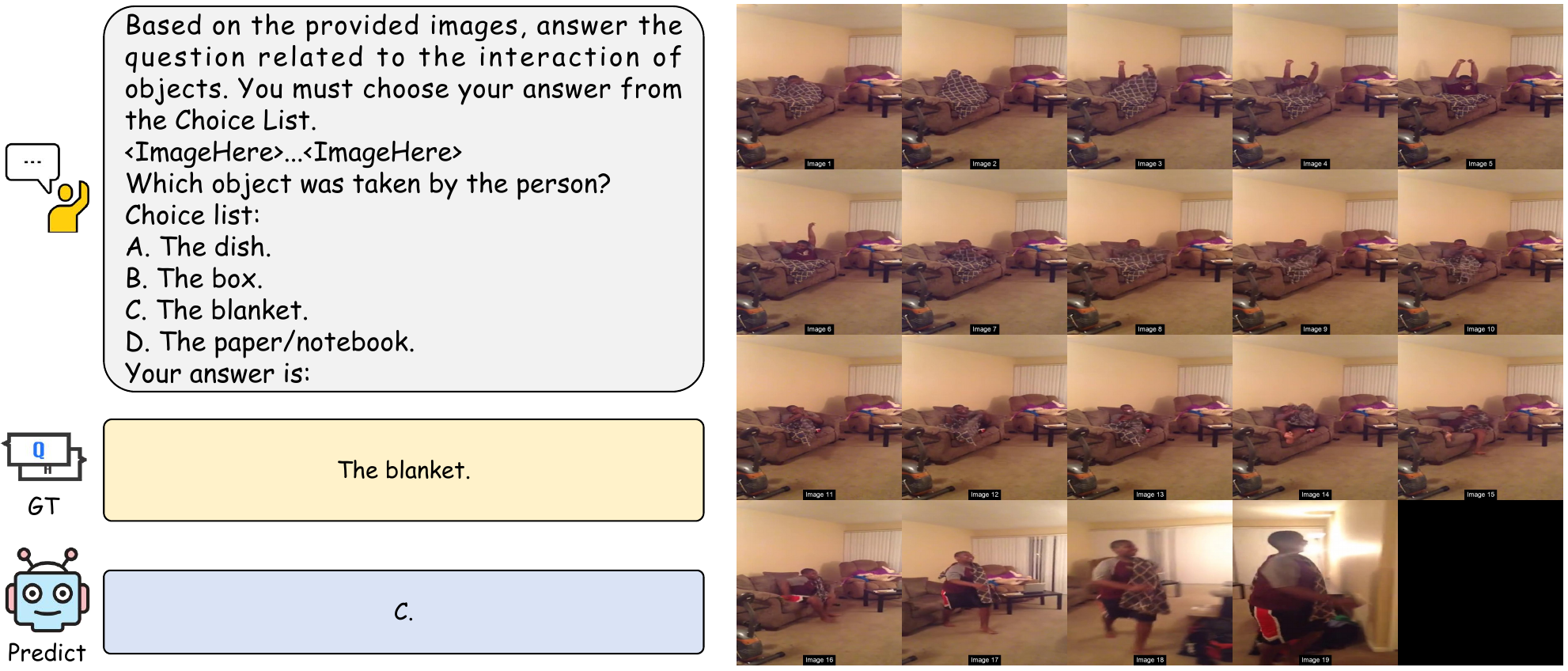}
        \caption{}
    \end{subfigure}

    \begin{subfigure}{\linewidth}
    \centering
    \includegraphics[width=0.75\textwidth]{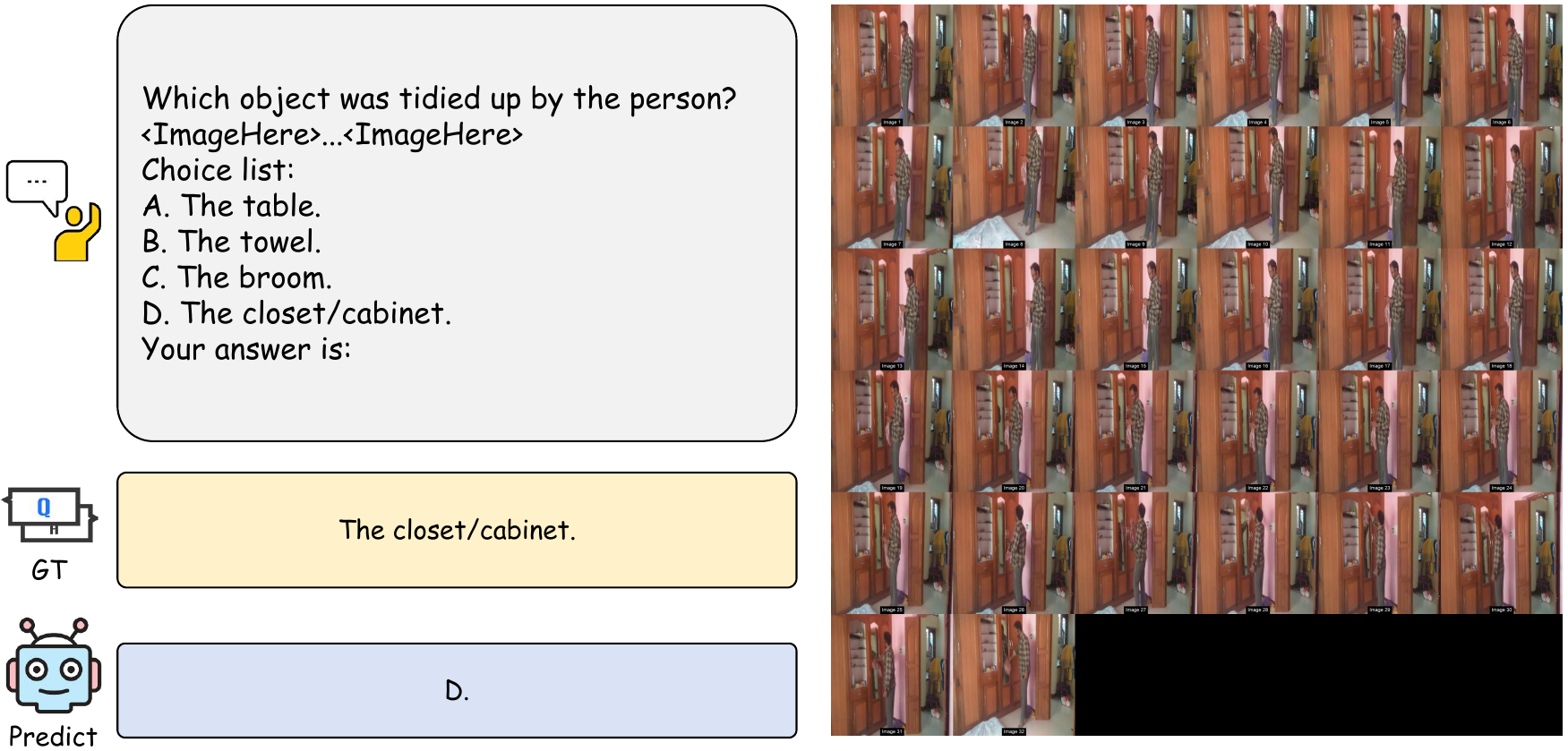}        
    \caption{}
    \end{subfigure}
    \caption{Case study of our pruning strategy on several LLaVA-1.5-7B samples. GT denotes Ground Truth.}
    \label{fig:case_study}
\end{figure*}

To qualitatively evaluate the effectiveness of our pruning strategy, we conduct a case study on several representative samples using LLaVA-1.5-7B. Figure~\ref{fig:case_study} presents illustrative examples that highlight the adaptive behavior of our method. For sequential tasks involving object interactions, our approach leverages image variations to effectively filter redundant information from static backgrounds, thereby allocating the limited token budget to dynamic foreground entities. As a result, the model can accurately identify critical objects even under highly compressed visual representations. In the query case related to the Scottish League Cup, the diversity metric ensures that nuanced, semantically important details such as text and logos are preserved. Overall, these results demonstrate that our method can effectively distinguish redundant visual content from semantically salient signals.

\end{document}